\documentclass[letterpaper]{article} 
\usepackage[preprint]{aaai2027}  
\usepackage[hyphens]{url}  
\usepackage{graphicx} 
\usepackage{natbib}  
\usepackage{caption} 
\usepackage{algorithm}
\usepackage{algorithmic}

\usepackage{newfloat}
\usepackage{listings}
\DeclareCaptionStyle{ruled}{labelfont=normalfont,labelsep=colon,strut=off} 
\floatstyle{ruled}
\newfloat{listing}{tb}{lst}{}
\floatname{listing}{Listing}

\usepackage{booktabs}
\usepackage{array}
\usepackage{tabularx}
\newcolumntype{Y}{>{\raggedright\arraybackslash}X}

\title{
    How Do Large Language Models Judge Social Attraction?\\
    Evidence from Theory-Grounded Persona Ratings Across Multiple LLMs\\
    and Humans
}
\author{
    Hasan Mahmud\textsuperscript{\rm 1}\corresponding,
    Khawaja Abaid Ullah\textsuperscript{\rm 2},
    Mohammad Javad Khojasteh\textsuperscript{\rm 2},\\
    Jamison Heard\textsuperscript{\rm 2},
    Prabu David\textsuperscript{\rm 1}
}

\affiliations{
    \textsuperscript{\rm 1}School of Communication,
    Rochester Institute of Technology\\
    \textsuperscript{\rm 2}Department of Electrical and Microelectronic Engineering,
    Kate Gleason College of Engineering,
    Rochester Institute of Technology\\
    Rochester, NY 14623, USA\\
    hxmdfp@rit.edu,
    ka4150@rit.edu,
    mjkeme@rit.edu,
    jrheee@rit.edu,
    pxdpro@rit.edu
}

\begin{document}

\maketitle

\begin{abstract}
Large language models (LLMs) are increasingly used to perform subjective evaluations traditionally made by humans, yet their validity as social judges remains unclear. This paper examines whether LLMs can assess social attraction from theory-grounded persona profiles constructed from ten psychological and relational constructs and organized into three tiers: socially attractive, socially mixed, and socially unattractive. We examine LLM ratings in two studies and compare them with human judgments in a third study. In Study 1, 34 LLMs rated 12 profiles across three repeated runs. Although some models tended to give higher or lower ratings overall, they showed strong stability across runs, consistent three-tier ordering, and high agreement in relative profile ordering. Study 2 examined sensitivity to gender presentation using six matched name-and-pronoun profile pairs and a separate pronoun-only test with a gender-neutral name, finding no significant effects in either analysis. In Study 3, 198 human participants evaluated the six matched profiles from Study 2. Their ratings reproduced the three-tier structure and followed a profile ordering consistent with that of the LLMs. However, LLMs rated attractive profiles more positively and unattractive profiles more negatively than humans, while neither group showed a significant overall effect of gender presentation.
\end{abstract}

\begin{links}
    \link{Code and data}{https://github.com/the-kas-lab/llm-social-attraction}
\end{links}

\section{Introduction}

Large language models (LLMs) are increasingly used to make judgments about people. They generate, interpret, and evaluate text representing human beliefs, preferences, behaviors, and social situations. Foundational work showed that large-scale language models can perform diverse tasks from instructions or examples, making them relevant to textual judgment and classification~\cite{brown2020language}. Recent studies have examined LLMs for survey research, experiments, automated content analysis, and simulations of human behavior~\cite{aher2023using,bail2024can,park2023generative}. These developments raise an important question: how should LLMs be evaluated when they make subjective social judgments rather than factual or technical assessments? Such judgments often require interpreting incomplete information, integrating multiple interpersonal cues, and evaluating constructs without a single correct answer or ground truth. In these settings, fluent responses do not necessarily indicate valid judgment, and agreement among models does not necessarily imply alignment with human evaluations.

A common form of social judgment concerns whether others seem likable, approachable, easy to interact with, and desirable. These evaluations reflect social attraction, one dimension of interpersonal attraction alongside task and physical attraction~\cite{mccroskey1974measurement}. Physical attraction concerns a person’s appearance, task attraction concerns their desirability as a work or problem-solving partner, and social attraction focuses on friendship potential and interpersonal fit. Because social-attraction judgments require integrating multiple interpersonal cues without an objectively correct answer, they provide a useful setting for examining LLM evaluations of socially meaningful information about people.
LLMs offer both opportunities and challenges for studying social judgment. They can repeatedly evaluate profiles under common instructions and reveal cross-model consistency. However, because LLMs are trained on large text corpora, their evaluations may reflect stereotypes, cultural assumptions, and social biases~\cite{bender2021dangers,navigli2023biases}. Their judgments may therefore appear coherent while reproducing dominant cultural norms, diverging from human evaluations, or exhibiting model-specific rating tendencies. Prior research has likewise shown that LLMs can reproduce some human-subject findings while distorting others~\cite{aher2023using}. Evaluating LLM social judgment therefore requires assessing reliability across repeated evaluations, convergence across models, sensitivity to theoretically relevant cues, and alignment with human ratings.

We examined these issues using theory-grounded persona profiles that varied across ten interpersonal constructs: attachment orientation, warmth, emotional intelligence, self-awareness, authenticity, conflict style, reciprocity, narcissistic entitlement, relational aggression, and emotional dysregulation. Different configurations of these characteristics were used to create profiles organized into three theoretically defined tiers: socially attractive, socially mixed, and socially unattractive. 
Study 1 used 34 LLMs to rate 12 profiles across three repeated runs, examining reliability, cross-model agreement, tier differentiation, and variation in rating patterns. Study 2 examined sensitivity to gender presentation using six matched male- and female-presented profile pairs and a separate pronoun-only test with a gender-neutral name presented with he/him, she/her, and they/them pronouns. Study 3 asked 198 human participants to evaluate the same six matched profile pairs to assess the three-tier structure, gender-presentation effects, and alignment between human and LLM judgments.

We addressed three research questions (RQs): \emph{RQ1}: Do LLMs reliably distinguish among theoretically defined levels of social attraction? \emph{RQ2}: To what extent do social-attraction judgments converge across LLMs and align with human ratings? \emph{RQ3}: Are social-attraction rating patterns influenced by gender presentation through names and pronouns?
This study contributes to research on LLM evaluation and human-centered AI in three ways. First, it introduces social attraction as a theory-grounded setting for evaluating subjective interpersonal judgment. By using controlled profiles that vary in expected social attractiveness, the study extends LLM evaluation beyond factual and task-based benchmarks to a setting requiring the integration of socially meaningful cues. Second, it provides a multi-model evaluation of 34 LLMs, examining repeated-run reliability, tier differentiation, cross-model agreement, and variation in rating patterns. Third, it compares aggregate LLM judgments with human ratings, providing evidence about outcome-level alignment between synthetic and human raters while also examining whether both groups’ judgments remain stable across changes in gender presentation.

\section{Theoretical Background}

\subsection{Social Attraction as a Judgment Task}

Social attraction is one dimension of interpersonal attraction, distinct from physical attraction and task attraction~\cite{mccroskey1974measurement}. It concerns whether a profile signals friendship potential, ease of interaction, social fit, and likability. A profile does not contain a definitive social-attraction label. Instead, raters infer how rewarding, safe, and sustainable interaction with that person might be. Positive cues, such as warmth, reciprocity, emotional understanding, and stable conflict behavior, may reinforce one another, but their effects may be weakened by inconsistency, entitlement, emotional volatility, or aggression.

This configurational character makes social attraction a useful setting for evaluating algorithmic judgment. Unlike an objective benchmark, the task requires graded interpretation of socially rich text. It can nevertheless be structured using profiles that vary in theoretically relevant cues. Profiles with strong warmth, reciprocity, emotional understanding, and stable conflict behavior should generally appear more attractive than profiles marked by contempt, relational aggression, entitlement, or chronic dysregulation. Profiles combining positive and negative relational cues should receive intermediate ratings.

\subsection{LLMs as Social Judgment Systems}

LLMs learn from human-generated text containing descriptions of people, relationships, stereotypes, and social norms, which may support impression formation and interpersonal evaluation~\cite{demszky2023using}. Prior work suggests that LLMs can approximate some human judgments in social and psychological tasks, raising the possibility that they may function as synthetic evaluators in selected research settings~\cite{dillion2023can,mittelstadt2024large}. However, several concerns motivate the present study. First, the alignment between models and people is uneven across tasks: models can flatten the diversity of human responses, inherit culturally skewed social norms, and overstate or understate effects relative to human samples~\cite{lin2025six,naous2024having,tao2024cultural,wang2025large}. Second, different LLMs are trained on different data and use different alignment procedures, which may lead them to diverge even when each produces individually plausible judgments.~\citet{qian2026can} showed that LLM judges vary in reliability and may produce inconsistent comparative assessments, cautioning against treating models as equally dependable or relying on unweighted aggregation. Accordingly, we examined repeated-run reliability and cross-model agreement across multiple LLMs and compared their aggregate ratings with a human benchmark.

\subsection{Gender Presentation and Bias in LLM Social Judgment}
Gendered names and pronouns may affect LLM judgments even when profile content remains unchanged. Because LLMs learn from human-generated text, their evaluations may reflect gender-linked expectations about warmth, competence, assertiveness, emotional responsiveness, and relational behavior. \citet{levy2024gender} found that LLM decisions differed across male, female, and gender-neutral names. \citet{an2024large} found that hiring decisions varied across names associated with gender, race, and ethnicity, partly depending on prompt wording. \citet{kotek2023gender} showed that occupational judgments often reflected gender stereotypes. \citet{hossain2023misgendered} found that language models performed worse with gender-neutral than binary pronouns and exhibited memorized name-pronoun associations. Together, these findings motivate a controlled matched name-and-pronoun manipulation and a pronoun-only comparison using a gender-neutral name.

\section{Empirical Studies}

The design comprised three studies examining LLM reliability, profile differentiation, sensitivity to gender presentation through names and pronouns, and correspondence with human judgments. Table 1 summarizes the raters, stimuli, designs, and primary analyses across the three studies.

\begin{table*}[t]
\centering
\begin{tabularx}{\textwidth}{
    @{}
    c
    >{\raggedright\arraybackslash}p{0.13\textwidth}
    Y
    Y
    @{}
}
\hline
\textbf{Study} &
\textbf{Raters} &
\textbf{Stimuli and design} &
\textbf{Primary evaluations} \\
\hline

1 & \begin{tabular}[t]{@{}l@{}}
34 LLMs\\
Three runs
\end{tabular} &
Twelve undergraduate personas: four socially attractive, four socially
mixed, and four socially unattractive. &
Repeated-run stability, tier differentiation, cross-model agreement,
and variation in ratings across models. \\

2 & \begin{tabular}[t]{@{}l@{}}
14 LLMs\\
Three runs
\end{tabular} &
Six employee-profile pairs with Original and Gender-swapped
presentations, created by manipulating names and pronouns; one
additional Alex Wilson profile presented with he/him, she/her, and
they/them pronouns. &
Profile-tier differentiation, sensitivity to gender presentation,
and the controlled pronoun-only test. \\

3 &
198 human participants &
The same six employee-profile pairs used in Study~2; between-subjects
assignment to either the Original or Gender-swapped presentation. &
Human-rating reliability, tier differentiation, sensitivity to gender
presentation, and matched human--LLM alignment. \\

\hline
\end{tabularx}
\caption{Overview of the three studies.}
\label{tab:overview_studies}
\end{table*}

\subsection{Study 1: Social Attraction Evaluation by LLMs}

\subsubsection{Stimuli}

The stimuli consisted of 12 hypothetical undergraduate student profiles containing contextual information about academic year, major, college affiliation, personal background, and interests. The profiles represented three theoretically defined levels of expected social attraction: socially attractive, socially mixed, and socially unattractive. Each tier included four profiles, two presented as male and two as female.

The profiles were constructed from social and relational cues grounded in psychological and interpersonal theories. Attachment theory informed each persona’s orientation toward closeness, dependence, and relational threat~\cite{bowlby1969attachment,hazan1987romantic}. Warmth was grounded in interpersonal circumplex work, especially the communion dimension of interpersonal behavior~\cite{wiggins1979psychological}. Emotional intelligence and self-awareness captured the ability to perceive, understand, and regulate emotions and recognize their effects on others~\cite{duval1972theory,mayer1997emotional}. Authenticity reflected consistency between their values, self-knowledge, and behavior in relationships~\cite{wood2008authentic}. Conflict style represented responses to disagreement and interpersonal tension, while reciprocity represented patterns of giving, balancing, conditioning, or taking relational effort~\cite{rahim1983measure}. Negative relational cues included narcissistic entitlement, relational aggression, and emotional dysregulation, representing self-focused, harmful, or unstable patterns in social interaction~\cite{crick1995relational,gratz2004multidimensional,raskin1988principal}.

The research team developed the theory-grounded construct specifications, and Claude Sonnet 4.6, which was not a rating model, produced initial narrative drafts. All profiles were then manually reviewed and revised for construct fidelity, internal consistency, realism, and differentiation. The complete construct configurations are provided in Supplementary Table S1. The profile-development process also incorporated theoretical consistency constraints when combining the constructs. For example, high emotional intelligence required at least moderate self-awareness because it involves perceiving and regulating one’s emotions~\cite{mayer1997emotional}. Similarly, high authenticity required at least moderate emotional intelligence because authentic self-expression involves self-knowledge and alignment between internal experience and outward behavior~\cite{wood2008authentic}. These constraints helped avoid incoherent combinations, such as high emotional intelligence with very low self-awareness or high authenticity with little understanding of one’s own emotions.
The final stimuli were organized into three tiers. Socially attractive profiles included strong positive cues and no active negative cues, although they could contain limited non-destructive flaws such as conflict avoidance or mild analytic distance. Socially mixed profiles combined an interpersonal strength with a clear relational limitation and at least one moderate relational risk. Socially unattractive profiles included weaker positive cues and stronger negative cues, such as chronic dysregulation, narcissistic entitlement, overt relational aggression, or contemptuous disengagement. Full persona specifications, profile texts, prompts, model identifiers, and technical configurations are provided in Supplementary Files 1 and 2; deidentified ratings and detailed analysis outputs are provided in Supplementary Files 3 and 4.

\subsubsection{LLM raters and rating procedure}
Thirty-four LLMs spanning multiple providers and model families were used as synthetic raters, including providers such as OpenAI, Google, Anthropic, xAI, NVIDIA, and Amazon, and model families such as Llama, Qwen, DeepSeek, Kimi, MiniMax, and MiMo. Specific model labels, providers, and OpenRouter identifiers are reported in Supplementary File 2. This multi-model design enabled examination of variation in rating patterns, test–retest reliability, and scale use across contemporary LLMs.

Each model was instructed to act as an independent rater and to evaluate each profile using only the information provided. Each of the 12 persona profiles was rated across three runs, with every profile evaluation submitted through a separate API call. For each profile, models completed a five-item social-attraction scale adapted from \citet{mccroskey1974measurement}. Responses were recorded on a seven-point Likert scale ranging from 1 (strongly disagree) to 7 (strongly agree), with Items 2 and 4 reverse-scored. The five scored items were averaged to produce a social-attraction score ranging from 1 to 7. The three runs were retained separately for the test–retest reliability analysis. For the tier-level, profile-level, and cross-model analyses, ratings were averaged across runs within each model and profile.

The rating pipeline was implemented through the OpenRouter API using a common prompt template across all models, with temperature fixed at 0.7 across all models and runs to maintain a consistent sampling configuration. For each model-profile-run combination, the dataset recorded the model identifier, profile identifier, run number, randomized execution order, and five item-level responses.

\subsubsection{Results}

\paragraph{Test-Retest Reliability Across Runs}
The repeated-run design allowed us to assess the stability of each model’s ratings. Across the 34 models, the mean single-measure absolute-agreement ICC was $.965$, with a median of $.970$, indicating high overall test–retest reliability. Twenty-eight of the 34 models had ICC values of at least $.95.$ Model-specific ICC estimates are reported in the “Reliability Summary” sheet of Supplementary File 4.

These results indicate that most models produced highly stable ratings across repeated evaluations of the same persona profiles. Reliability nevertheless varied across models, with some showing greater run-to-run variation than others. This variation highlights the need to evaluate LLMs not only by their mean ratings but also by their consistency as synthetic raters.

\paragraph{Tier Differentiation and Profile-Level Ratings}

The LLM ratings reproduced the expected tier-level ordering. Socially attractive profiles received the highest ratings ($M = 6.16$, $SD = .26$), followed by socially mixed profiles ($M = 4.70$, $SD = .34$) and socially unattractive profiles ($M = 2.93$, $SD = .41$). A one-way repeated-measures ANOVA of model-level tier means, with profile tier as the within-model factor, showed a strong effect of profile tier, $F(2,66)=800.94$, $p<.001$, $\eta_p^2=.960$. Follow-up paired-samples t-tests showed that all three tier contrasts were significant, all $p<.001$.

For each of the 34 models, the average rating for socially attractive profiles exceeded the average for socially mixed profiles, which in turn exceeded the average for socially unattractive profiles. Profile means largely followed the intended tier ordering, with one cross-tier crossover: Courtney Briggs, a socially unattractive profile, received a higher mean rating than Aaron Kowalski, a socially mixed profile. Dominic Reyes received the highest mean rating overall ($M=6.47$, $SD=.34$), whereas Natalie Voss received the lowest ($M=2.05$, $SD=.44$). Profile-level ratings are presented in Figure~\ref{fig:ratings_by_profile}, with detailed estimates reported in the supplementary materials.

\begin{figure}[t]
\centering
\includegraphics[width=\columnwidth]{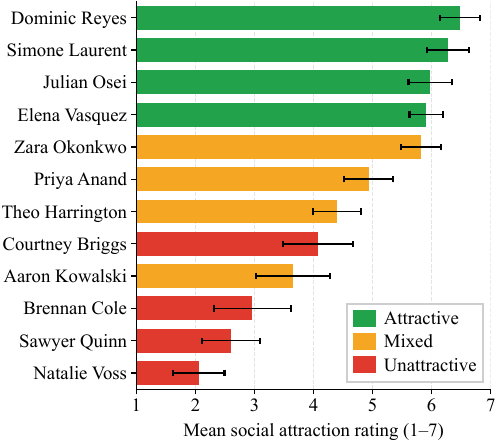}
\caption{Mean social-attraction ratings for the 12 persona profiles across 34 LLMs and three repeated runs. Bars are color-coded by intended profile tier: socially attractive, socially mixed, and socially unattractive. Error bars indicate $\pm1$ SD across the 34 model-level profile means. }
\label{fig:ratings_by_profile}
\end{figure}

\paragraph{Cross-Model Agreement and Rating Variation}
Agreement across the 34 LLMs was high. The models showed strong consistency in their ordering of the 12 profiles, Kendall’s $W=.948$, $\chi^2 (11)=354.47$, $p<.001$, and the mean pairwise Spearman correlation was .951. However, agreement in profile ordering did not mean that all models used the rating scale identically. Model-level means ranged from $3.87$ to $5.09$, indicating differences in overall rating leniency, while within-model standard deviations ranged from $1.10$ to $2.41$, indicating differences in how strongly models distinguished among profiles. Variation was lowest for clearly attractive profiles and greater for mixed and unattractive profiles. Detailed model-level estimates are provided in Supplementary File 4, and the relationship between test–retest reliability and profile differentiation is shown in Supplementary Figure S2.

\subsection{Study 2: Sensitivity to Gender Presentation}

Because names and pronouns can activate gender-linked expectations, Study 2 examined whether LLM social-attraction judgments changed when gender presentation through names and pronouns was altered while the underlying relational content remained constant. In addition to the matched gender-presentation analysis that changed both names and pronouns, a pronoun-only analysis using the gender-neutral name Alex Wilson was added.

\subsubsection{Matched Gendered Name-and-Pronoun Test}

Six personas were selected from Study 1, with two profiles from each social-attraction tier and balanced male and female presentations. The original undergraduate personas were adapted into shorter employee profiles situated in an AI startup and used to create six matched pairs in which profile content was held constant while names and pronouns were changed to alter gender presentation. For example, as shown in Table 2, Dominic Reyes became Dominique Reyes, and pronouns changed from he/him/his to she/her/hers.

Limiting the stimulus set to six profiles kept Study 3 manageable while preserving all three tiers and enabling direct human–LLM comparison. Student-specific details were removed to suit a heterogeneous adult sample and reduce reading burden while preserving each persona’s central social and relational characteristics and intended tier. Fourteen LLMs were selected to represent a range of frontier and widely used models. The set spanned providers such as OpenAI, Google, Anthropic, xAI, Meta, Alibaba, Amazon, and Xiaomi, and included model families such as Llama, Qwen, Nova, MiMo, and GPT-OSS. Each model evaluated the Original and Gender-swapped versions of the six matched pairs across three runs. Ratings were averaged across runs within each model, profile, and gender-presentation condition. Full profile texts, prompts, model labels, providers, OpenRouter identifiers, technical configurations, ratings, and analysis outputs are provided in Supplementary Files 5 and 6.

\begin{table*}[t]
\centering
\begin{tabularx}{\textwidth}{@{}YY@{}}
\hline
\textbf{Original presentation} &
\textbf{Gender-swapped presentation} \\
\hline

\textbf{Dominic Reyes:}
Dominic is good at reading people fast. [\ldots]
\textbf{He} is the kind of person who makes a group work. [\ldots]
\textbf{His} tension point is conflict, and he knows it. [\ldots]
By the time \textbf{he} admits something is wrong, it has usually
been wrong for a while.
&
\textbf{Dominique Reyes:}
Dominique is good at reading people fast. [\ldots]
\textbf{She} is the kind of person who makes a group work. [\ldots]
\textbf{Her} tension point is conflict, and she knows it. [\ldots]
By the time \textbf{she} admits something is wrong, it has usually
been wrong for a while.
\\

\hline
\end{tabularx}
\caption{Example of an original profile and its gender-swapped presentation.
Boldface indicates the features modified between presentations.}
\label{tab:gender_swapped_example}
\end{table*}

\paragraph{Results}
The three-tier pattern remained clear in the Gender-swapped condition: socially attractive profiles received the highest ratings ($M=6.23$, $SD=.41$), followed by socially mixed ($M=4.18$, $SD=.76$) and socially unattractive profiles ($M=2.22$, $SD=.55$). A $2 \times 3$ repeated-measures ANOVA found a significant effect of profile tier, $F(2,26)=276.91$, $p<.001$, $\eta_p^2=.955$, but no effect of gender presentation, $F(1,13)=0.03$, $p=.876$, $\eta_p^2=.002$, and no interaction, $F(2,26)=1.09$, $p=.350$, $\eta_p^2=.078$. Overall ratings were nearly identical for the Original ($M=4.22$) and Gender-swapped profiles ($M=4.21$), and rank-order agreement across the six profiles was strong, Spearman’s $\rho=.943$, $p=.005$.

\subsubsection{Controlled Pronoun-Presentation Test Including a Gender-Neutral Condition}
Because the matched analysis changed both names and pronouns and included only male and female presentations, we conducted a controlled pronoun-only test using Alex Wilson, a socially mixed profile with a gender-neutral name. The profile was presented with he/him, she/her, or they/them pronouns while the name and relational content remained unchanged. Fourteen LLMs evaluated all three versions across three runs, and scores were averaged within model and pronoun presentation before analysis. Full materials and results are provided in Supplementary Files 10 and 11.

\paragraph{Results}
Ratings were similar across the male ($M=5.39$, $SD=.61$), female ($M=5.28$, $SD=.59$), and gender-neutral presentations ($M=5.32$, $SD=.62$), and the effect of pronoun presentation was not significant, $F(2,26)=1.52$, $p=.238$, $\eta_p^2=.105$. Holm-adjusted pairwise comparisons were nonsignificant, and the planned comparison between the gender-neutral presentation and the average of the two gendered presentations was also nonsignificant, $t(13)=-0.16$, $p=.873$, $d_z=-.04$.

\subsection{Study 3: Human Validation}

Study 3 introduced a human benchmark to examine whether human social-attraction judgments aligned with aggregate LLM ratings and remained stable across gender presentation.

\subsubsection{Design and Participants}

Human participants evaluated the same six employee profiles from Study 2, presented in either their Original or Gender-swapped form, using the same five-item social-attraction measure. Participants were randomly assigned to one of the two conditions. The recruitment target of 200 was intended to provide approximately 100 ratings per profile in each condition. A sensitivity analysis based on the final condition sizes of 97 and 101 participants indicated that a two-sided between-condition comparison at $\alpha=.05$ had $80\%$ power to detect Cohen’s $d=0.40$.

Participants were recruited through Prolific and completed the study through Qualtrics. The study was determined exempt by the Human Subjects Research Office, and all participants provided informed consent. Of the 200 participants, two failed the attention check and were excluded, resulting in N = 198: 97 in the Original condition and 101 in the Gender-swapped condition. Participants were aged 19--69 years ($M=31.39$, $SD=9.39$). Participants received £2.50, and the median completion time was 19 minutes 28 seconds, corresponding to £7.71 per hour.

Responses were recorded on a seven-point scale from 1 (strongly disagree) to 7 (strongly agree), with Items 2 and 4 reverse-scored. The survey also included an attention check and demographic questions. The complete survey instrument and deidentified data are provided in Supplementary Files 7–9.

\subsubsection{Results}

\paragraph{Scale Reliability and Inter-Rater Agreement}

The five-item social-attraction scale demonstrated high internal
consistency, Cronbach's $\alpha=.937$. Individual-rater agreement was moderate in the Original and Gender-swapped conditions, ICC(A,1) $=.557$ and $.532$, respectively, whereas the aggregated profile ratings were highly reliable, ICC(A,k) = $.992$ and $.991$, respectively. Complete estimates and confidence intervals are provided in Supplementary File 12.

\paragraph{Tier-Level Findings}

Human ratings followed the expected social-attraction structure. In the Original condition, socially attractive profiles received the highest ratings ($M=5.69$, $SD=.96$), followed by socially mixed ($M=4.53$, $SD=1.28$) and socially unattractive profiles ($M=2.72$, $SD=1.31$). In the Gender-swapped condition, socially attractive profiles also received the highest ratings ($M=5.57$, $SD=1.02$), followed by socially mixed ($M=4.31$, $SD=1.31$) and socially unattractive profiles ($M=2.79$, $SD=1.23$).

To determine whether participants distinguished among the three intended tiers, paired-samples t-tests were conducted. The results showed that in both conditions, attractive profiles were rated higher than mixed profiles, mixed profiles were rated higher than unattractive profiles, and attractive profiles were rated higher than unattractive profiles. All three pairwise contrasts were significant in the Original condition, with $t(96)$ values ranging from $8.91$ to $21.52$ and $d_z$ values ranging from $.91$ to $2.18$, and in the Gender-swapped condition, with t(100) values ranging from $12.10$ to $20.21$ and $d_z$ values ranging from $1.20$ to $2.01$. All p-values were below $.001$. Complete contrast-specific results are provided in Supplementary File 12.

\subsubsection{Sensitivity to Gender Presentation}
We examined whether human social-attraction ratings changed when identical persona content was presented with Original or Gender-swapped names and pronouns. A linear mixed-effects model included gender-presentation condition, profile tier, and their interaction as fixed effects, with a participant random intercept. Profile tier had a significant effect, likelihood-ratio $\chi^2 (2)=834.00$, $p<.001$. Neither gender presentation, $\chi^2 (1)=1.12$, $p=.290$, nor its interaction with profile tier, $\chi^2 (2)=3.28$, $p=.194$, was significant. Thus, human raters distinguished the three profile tiers, but changing gendered names and pronouns did not significantly alter the overall rating pattern.

\subsubsection{Human–LLM Alignment and Rating Differences}
To assess human–LLM alignment under matched stimulus conditions, we compared profile-level human mean ratings with corresponding aggregate ratings from the 14 LLMs in Study 2. Human participants and LLMs evaluated identical employee-profile texts using the same five-item social-attraction measure.

Profile-level alignment was strong in both conditions. In the Original condition, human and LLM means were highly correlated, Pearson’s $r=.981$, $95\%$ CI $[.835, .998]$, $p<.001$, with strong rank-order agreement, Spearman’s $\rho=.886$, $p=.019$. In the Gender-swapped condition, the corresponding associations were Pearson’s $r=.979$, $95\%$ CI $[.812, .998]$, $p<.001$, and Spearman’s $\rho=.943$, $p=.005$.

Direct comparisons showed significant interactions between rater type and profile tier in both the Original condition, $\chi^2 (2)=20.36$, $p<.001$, and the Gender-swapped condition, $\chi^2 (2)=34.60$, $p<.001$. Holm-adjusted contrasts showed that LLMs rated attractive profiles more positively than humans in the Original ($\Delta=0.49$, $p<.001$) and Gender-swapped ($\Delta=0.66$, $p<.001$) conditions. LLMs also rated unattractive profiles more negatively in the Original ($\Delta=-0.49$, $p=.018$) and Gender-swapped ($\Delta=-0.57$, $p=.003$) conditions. Differences for the mixed tier were not significant in either condition (Original: $p=.097$; Gender-swapped: $p=.488$). Profile-specific comparisons further showed that LLMs rated Theo/Thea less favorably than humans in both the Original and Gender-swapped conditions, with Holm-adjusted $p=.003$ and $p=.040$, respectively.  Overall, humans and LLMs showed strong agreement in profile ordering, but LLMs differentiated attractive and unattractive profiles more strongly than humans, as illustrated in Figure 2.

\begin{figure}[t]
\centering
\includegraphics[width=\columnwidth]
{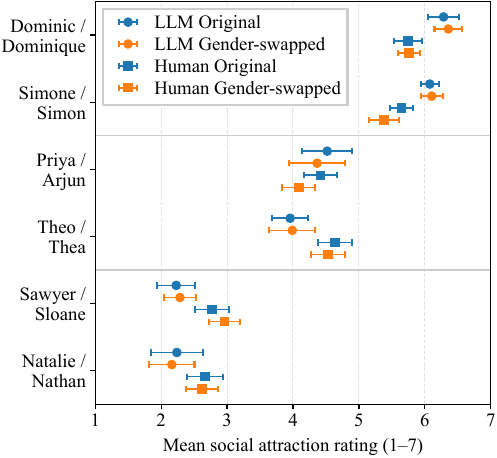}
\caption{Mean human and LLM social-attraction ratings for the six matched Original and Gender-swapped profile pairs. Error bars indicate 95\% confidence intervals within each rater type and gender-presentation condition.}
\label{fig:human_llm_ratings_comparison}
\end{figure}

\section{Discussion}

This study addressed three questions concerning the reliability, convergence, and gender sensitivity of LLM social-attraction judgments. Regarding RQ1, ratings across 34 LLMs were highly stable across repeated runs and consistently distinguished the three intended tiers. Regarding RQ2, the models showed strong cross-model agreement, and aggregate LLM ratings showed strong rank-order alignment with human ratings, although LLMs differentiated attractive and unattractive profiles more strongly. Regarding RQ3, neither LLM nor human ratings showed a significant overall effect of gender presentation. Overall, LLM rating patterns broadly resembled human ratings, consistent with prior evidence that LLMs can reproduce selected human-subject findings in controlled settings~\cite{aher2023using,argyle2023out}.

A central contribution of this study is the distinction between alignment in profile ordering and rating scale use. Humans and LLMs generally agreed on which profiles were more or less socially attractive, yet LLMs differentiated attractive and unattractive profiles more strongly. LLMs may therefore capture the relative social meaning of profile configurations while applying a more contrastive scale than human raters. LLMs may be useful for comparing standardized profiles, but their numerical ratings should not be treated as interchangeable with human judgments.

Gender-related biases have been documented across AI systems, including commercial applications~\cite{raji2019actionable}. Similar concerns extend to LLMs because their training data may encode social biases~\cite{bender2021dangers}. Empirical research has shown that LLMs can make gender-stereotyped judgments~\cite{kotek2023gender} and generate stereotyped portrayals of demographic groups~\cite{cheng2023marked}. Against this background, the gender-swap analyses provided a controlled test. Because the matched profiles preserved relational content and differed only in names and pronouns, significant differences would have suggested sensitivity to gender presentation rather than profile content. Instead, neither LLM nor human ratings showed a significant overall effect of gender presentation. The pronoun-only comparison using the gender-neutral name Alex Wilson also held name and content constant while varying he/him, she/her, and they/them pronouns. Ratings again did not differ significantly.

These findings have implications for using LLMs as synthetic raters in social-judgment research. At the profile level, humans and LLMs ordered most profiles similarly, but the clearest divergence involved Theo/Thea. In the Original condition, humans ranked Theo above Priya, whereas LLMs ranked Priya above Theo. The pattern repeated in the Gender-swapped condition: humans ranked Thea above Arjun, whereas LLMs ranked Arjun above Thea. Theo/Thea’s limited emotional warmth and inconsistent availability may have led LLMs to rate the profile less favorably than humans. In Study 1, Courtney Briggs also received a higher rating than Aaron Kowalski despite belonging to a lower intended tier, possibly because her social fluency and humor offset some relational risks. These interpretations remain tentative because individual profile cues were not experimentally isolated. Models also differed in overall rating levels and how strongly they differentiated among profiles, consistent with broader evidence of cross-model variation~\cite{liang2023holistic}. Together, these differences show that aggregate alignment can conceal important variation. Researchers should therefore compare multiple models and use human benchmarks. LLMs should be treated as complementary research instruments rather than replacements for human participants, particularly when social identity and lived experience are relevant~\cite{wang2025large}.

The findings also raise ethical and methodological issues. Social-attraction judgments involve evaluative assessments of people, and such judgments may carry consequences if used in systems for hiring, team formation, matchmaking, education, or social recommendation.  Human–LLM alignment does not establish fairness, appropriateness, or safety because human judgments may also reflect cultural assumptions and social biases. Research has shown that the opinions expressed by LLMs may poorly represent some demographic groups and that using models as substitutes for human populations can flatten differences among identity groups~\cite{santurkar2023whose,wang2025large}. LLMs may also reproduce majority norms, cultural expectations, or hidden stereotypes, especially in less controlled settings. These risks are consistent with broader taxonomies identifying discrimination, exclusion, manipulation, and human–computer interaction harms associated with language models~\cite{weidinger2022taxonomy}. Therefore, LLM-based social judgment tools should be designed with human oversight, transparent evaluation procedures, and clear limits on use. Human-in-the-loop evaluation remains important, particularly when judgments concern social suitability, interpersonal fit, or other subjective assessments of people.

Overall, human--LLM alignment should be interpreted as similarity in judgment patterns, not equivalence in fairness, appropriateness, or suitability for real-world use.

\section{Limitations and Future Research}
This study used researcher-designed profiles with LLM-assisted narrative drafting, which enabled controlled comparisons but may have presented social cues more clearly than naturally occurring introductions. Future research should examine more ambiguous, naturalistic, multimodal, and dialogue-based stimuli that capture tone, nonverbal cues, and repeated interaction. The direct human–LLM comparison was limited to six persona identities, 14 LLMs, and two gender-presentation conditions. The small, theoretically structured stimulus set may have contributed to the strong alignment, and results may vary across model versions, prompts, provider configurations, and human populations. Larger, more heterogeneous evaluations are needed. The study measured perceived social attraction rather than actual social behavior. Future work should examine behavioral outcomes such as willingness to communicate, partner selection, or continued interaction. Finally, the gender-presentation tests manipulated names and pronouns while holding profile content constant. This controlled manipulation may have constrained gender-linked effects and did not capture broader or intersectional identity cues. The absence of significant differences should therefore not be interpreted as evidence that humans or LLMs are generally free from gender bias.

\section{Conclusion}
This study shows that LLMs can reflect broad patterns in human social-attraction judgments from controlled persona profiles. Across models, ratings consistently distinguished socially attractive, socially mixed, and socially unattractive profiles, and aggregate LLM ratings closely followed human profile ordering. However, humans and LLMs did not use the rating scale identically, and some profile-level judgments differed.

LLMs may serve as scalable comparative instruments for examining standardized social stimuli. Their use requires repeated evaluations, comparisons across models, and validation against human judgments because aggregate alignment can conceal differences in scale use and profile-specific judgments. The absence of significant gender-presentation effects applies only to the controlled manipulation of names and pronouns and should not be generalized to broader forms of gender bias. LLMs should therefore complement, rather than replace, human judgment in social-evaluation research.

\bibliography{references}


\end{document}